\pdfoutput=1

\documentclass[11pt]{article}

\usepackage{ACL2023}

\usepackage{times}
\usepackage{latexsym}

\usepackage[T1]{fontenc}

\usepackage[utf8]{inputenc}
\usepackage[russian,english]{babel}

\usepackage{microtype}

\usepackage{inconsolata}

\usepackage{hyperref}
\usepackage{url}
\usepackage{amssymb}
\usepackage{amsmath}
\usepackage{subfig}
\usepackage{graphicx}
\usepackage{graphics}
\usepackage{booktabs}
\usepackage{multirow}
\usepackage{color}
\usepackage{enumitem}
\usepackage{makecell}
\usepackage{comment}
\usepackage{nicefrac}
\usepackage{amsfonts}
\usepackage{bm}
\usepackage{amsthm}

\usepackage{enumitem}
\setlist[itemize,enumerate]{leftmargin=*}
\usepackage[normalem]{ulem}
\usepackage{pgfplots}
\usepackage{tikz}
\usepackage{xcolor,colortbl}
\usepackage{color,soul}
\usepackage{multirow}
\usepackage{comment}
\usepackage{xspace}
\usepackage{arydshln}
\usepackage{dsfont}

\makeatletter
\def\adl@drawiv#1#2#3{%
        \hskip.5\tabcolsep
        \xleaders#3{#2.5\@tempdimb #1{1}#2.5\@tempdimb}%
                #2\z@ plus1fil minus1fil\relax
        \hskip.5\tabcolsep}
\newcommand{\cdashlinelr}[1]{%
  \noalign{\vskip 1.3pt
           \global\let\@dashdrawstore\adl@draw
           \global\let\adl@draw\adl@drawiv}
  \cdashline{#1}[.4pt/2pt]
  \noalign{\global\let\adl@draw\@dashdrawstore
           \vskip 1.3pt}}
\makeatother

\newcommand{\norm}[1]{\left\lVert #1 \right\rVert}

\newcommand{\crestrat}{CREST-Rationalization\xspace}
\newcommand{\crestgen}{CREST-Generation\xspace}
\newcommand{\crest}{CREST\xspace}

\newcommand{\tquad}{\;\;\,}

\definecolor{seaborn-blue}{RGB}{76, 114, 176}
\definecolor{seaborn-orange}{RGB}{221, 132, 81}
\definecolor{seaborn-green}{RGB}{85, 168, 105}
\definecolor{seaborn-red}{RGB}{197, 79, 83}
\definecolor{seaborn-purple}{RGB}{129, 115, 180}
\definecolor{seaborn-brown}{RGB}{147, 120, 96}
\definecolor{seaborn-pink}{RGB}{218, 139, 195}
\definecolor{seaborn-gray}{RGB}{140, 140, 140}
\definecolor{seaborn-yellow}{RGB}{203, 185, 116}
\definecolor{seaborn-cyan}{RGB}{100, 180, 20}
\definecolor{paired-light-blue}{RGB}{198, 219, 239}
\definecolor{paired-dark-blue}{RGB}{49, 130, 188}
\definecolor{paired-light-orange}{RGB}{251, 208, 162}
\definecolor{paired-dark-orange}{RGB}{230, 85, 12}
\definecolor{paired-light-green}{RGB}{199, 233, 193}
\definecolor{paired-dark-green}{RGB}{49, 163, 83}
\definecolor{paired-light-purple}{RGB}{218, 218, 235}
\definecolor{paired-dark-purple}{RGB}{117, 107, 176}
\definecolor{paired-light-gray}{RGB}{217, 217, 217}
\definecolor{paired-dark-gray}{RGB}{99, 99, 99}
\definecolor{paired-light-pink}{RGB}{222, 158, 214}
\definecolor{paired-dark-pink}{RGB}{123, 65, 115}
\definecolor{paired-light-red}{RGB}{231, 150, 156}
\definecolor{paired-dark-red}{RGB}{131, 60, 56}
\definecolor{paired-light-yellow}{RGB}{231, 204, 149}
\definecolor{paired-dark-yellow}{RGB}{141, 109, 49}
\definecolor{light-green}{RGB}{118, 207, 180}
\definecolor{raspberry}{RGB}{228, 24, 99}
\definecolor{caddback}{rgb}{0.90, 0.98, 0.96}
\definecolor{cadd}{rgb}{0, 0.47, 0.34}
\definecolor{cdelback}{rgb}{1, 0.94, 0.92}
\definecolor{cdel}{rgb}{0.83, 0.32, 0.16}
\definecolor{ccon}{HTML}{fee9d4}
\definecolor{cood}{HTML}{d8f0d3}
\definecolor{cid}{HTML}{dae8f5}

\definecolor{set1-red}{HTML}{e41a1c}
\definecolor{set1-blue}{HTML}{377eb8}
\definecolor{set1-green}{HTML}{4daf4a}

\pgfplotsset{compat=1.9}

\pgfplotscreateplotcyclelist{my style}{%
    set1-red, solid, every mark/.append style={solid, fill=set1-red},mark size=1.5pt, mark=square*\\%
    set1-blue, dashed, every mark/.append style={solid, fill=set1-blue},mark size=1.5pt, mark=triangle*\\%
    set1-green!90!black, densely dotted, every mark/.append style={solid, fill=set1-green!90!black},mark size=1.5pt, mark=*\\%
}

\title{CREST: A Joint Framework for Rationalization and \\ Counterfactual Text Generation}

\author{
Marcos Treviso$^{1,2}$\thanks{~~Correspondence to: \texttt{marcos.treviso@tecnico.pt}}, 
Alexis Ross$^{3}$, 
Nuno M. Guerreiro$^{1,2}$, 
André F. T. Martins$^{1,2,4}$ \\
$^{1}$Instituto de Telecomunicações, Lisbon, Portugal \\ 
$^{2}$Instituto Superior Técnico \& LUMLIS (Lisbon ELLIS Unit), Lisbon, Portugal \\
$^{3}$Massachusetts Institute of Technology \\
$^{4}$Unbabel, Lisbon, Portugal
}

\begin{document}
\maketitle
\begin{abstract}
Selective rationales and counterfactual examples have emerged as two effective, complementary classes of interpretability methods for analyzing and training NLP models. However, prior work has not explored how these methods can be integrated to combine their complementary advantages. 
We overcome this limitation by introducing CREST (ContRastive Edits with Sparse raTionalization), a joint framework for selective rationalization and counterfactual text generation, and show that this framework leads to improvements in counterfactual quality, model robustness, and interpretability. 
First, CREST generates valid counterfactuals that are more natural than those produced by previous methods, and subsequently can be used for data augmentation at scale, reducing the need for human-generated examples. 
Second, we introduce a new loss function that leverages CREST counterfactuals to regularize selective rationales and show that this regularization improves both model robustness and rationale quality, compared to methods that do not leverage CREST counterfactuals. Our results demonstrate that CREST successfully bridges the gap between selective rationales and counterfactual examples, addressing the limitations of existing methods and providing a more comprehensive view of a model’s predictions.
\end{abstract}

\section{Introduction}

As NLP models have become larger and less transparent, there has been a growing interest in developing methods for finer-grained interpretation and control of their predictions. One class of methods leverages \textbf{selective rationalization}~\citep{lei-etal-2016-rationalizing,bastings-etal-2019-interpretable}, which trains models to first select \textit{rationales}, or subsets of relevant input tokens,
and then make predictions based only on the selected rationales. These methods offer increased interpretability, as well as learning benefits, such as improved robustness to input perturbations~\citep{jain-etal-2020-learning,chen-etal-2022-rationalization}. 
Another class of methods generates \textbf{counterfactual examples}, or modifications to input examples that change their labels. By providing localized views of decision boundaries, counterfactual examples can be used as explanations of model predictions, contrast datasets for fine-grained evaluation, or new training datapoints for learning more robust models \citep{ross-etal-2021-explaining,gardner-etal-2020-evaluating,Kaushik2020Learning}.

\begin{figure}[t]
    \centering
    \includegraphics[width=\columnwidth]{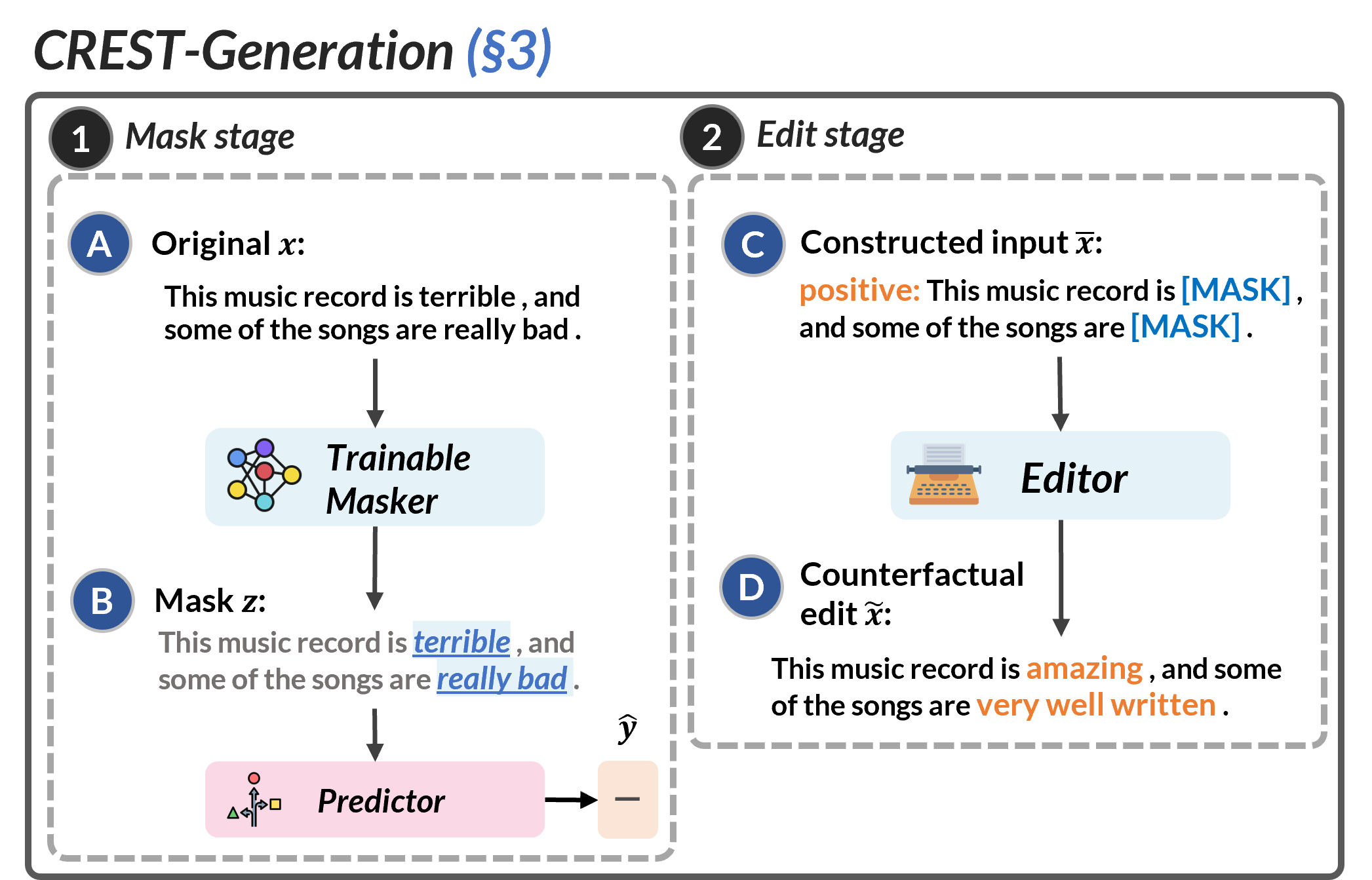}
    \caption{Our generation procedure consists of two stages: (i) 
    a mask stage that highlights relevant tokens in the input through a learnable masker; 
    and (ii) an edit stage, which receives a masked input and uses a masked language model to infill spans conditioned on a prepended label.}
    \label{fig:intro_fig}
\end{figure}

\begin{figure*}[t]
    \centering
    \includegraphics[width=1.0\linewidth]{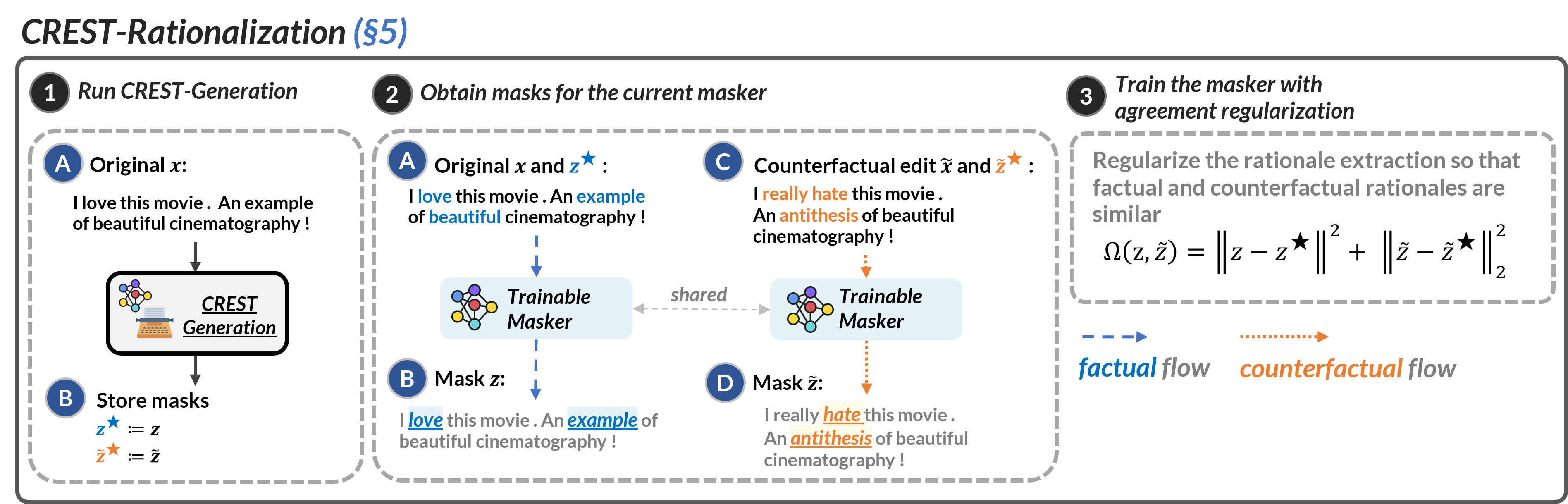}
    \caption{
    Overview of \crestrat. We start by passing an input $\bm{x}$ through \crestgen, which yields a counterfactual edit $\tilde{\bm{x}}$ along side two masks: $\bm{z}^\star$ for the original input, and $\tilde{\bm{z}}^\star$ for the counterfactual. 
    Next, we train a new rationalizer (masker) decomposed into two  flows: a \textbf{factual flow} that takes in $\bm{x}$ and produces a rationale $\bm{z}$, and a \textbf{counterfactual flow} that receives $\tilde{\bm{x}}$ and produces a rationale $\tilde{\bm{z}}$. Lastly, we employ a regularization term $\Omega(\bm{z}, \tilde{\bm{z}})$ to encourage agreement between rationales for original and counterfactual examples.
    }
    \label{fig:intro_fig_2}
\end{figure*}

This paper is motivated by the observation that selective rationales and counterfactual examples allow for interpreting and controlling model behavior through different means: selective rationalization improves model transparency by weaving interpretability into a model's internal decision-making process, while counterfactual examples provide external signal more closely aligned with human causal reasoning~\citep{wu-etal-2021-polyjuice}. 

We propose to combine both methods to leverage their complementary advantages. We introduce \textbf{\crest (\emph{ContRastive Edits with Sparse raTionalization})}, a joint framework for rationalization and counterfactual text generation.
\crest first generates high-quality counterfactuals (Figure~\ref{fig:intro_fig}), then leverages those counterfactuals to encourage consistency across ``flows'' for factual and counterfactual inputs (Figure~\ref{fig:intro_fig_2}). In doing so, \crest unifies two key important dimensions of interpretability introduced by \citet[\S3.2]{doshi2017towards}, forward simulation and counterfactual simulation. 
Our main contributions are:\footnote{Code at \url{https://github.com/deep-spin/crest/}.}

\begin{itemize}
    \item We present \textbf{\crestgen} (Figure~\ref{fig:intro_fig}), a novel approach to generating counterfactual examples by combining sparse rationalization with span-level masked language modeling (\S\ref{sec:crest_gen}),  
    which produces valid, fluent, and diverse counterfactuals (\S\ref{sec:counterfactual_generation}, Table~\ref{tab:results_counterfactuals_methods}).

    \item We introduce \textbf{\crestrat} (Figure~\ref{fig:intro_fig_2}), a novel approach to regularizing rationalizers. \crestrat decomposes a rationalizer into factual and counterfactual flows and encourages agreement between the rationales for both (\S\ref{sec:crest_rationalization}).

    \item We show that \crest-generated counterfactuals can be effectively used to increase model robustness, leading to larger improvements on contrast and out-of-domain datasets than using manual counterfactuals (\S\ref{subsec:results_training_with_counterfactuals}, Tables~\ref{tab:results_rationalizers_in_domain} and \ref{tab:results_rationalizers_in_domain_snli_hyp}). 
    
    \item We find that rationales trained with \crestrat not only are more plausible, but also achieve higher forward and counterfactual simulabilities
    (\S\ref{subsec:interpretability_analysis}, Table~\ref{tab:interpretability_analysis}).

\end{itemize}

Overall, our experiments show that \crest successfully combines the benefits of counterfactual examples and selective rationales to improve the quality of each, resulting in a more interpretable and robust learned model.

\section{Background}

\subsection{Rationalizers} \label{subsec:vanilla_rationalizer}

The traditional framework of rationalization involves training two components cooperatively: the \emph{generator}---which consists of an encoder and an explainer---and the \emph{predictor}. The generator encodes the input and produces a ``rationale'' (e.g., word highlights), while the predictor classifies the text given only the rationale as input~\citep{lei-etal-2016-rationalizing}. 

Assume a document $\bm{x}$ with $n$ tokens as input.
The encoder module (\textsf{enc}) converts the input tokens into $d$-dimensional hidden state vectors $\bm{H} \in \mathbb{R}^{n \times d}$, which are passed to the explainer (\textsf{expl}) to generate a latent mask $\bm{z} \in \{0, 1\}^n$. The latent mask serves as the rationale 
since it is used to select a subset of the input $\bm{x} \odot \bm{z}$, which is then passed to the predictor module (\textsf{pred}) to produce a final prediction $\hat{y} \in \mathcal{Y}$, where %
$\mathcal{Y} = \{1,...,k\}$ for $k$-class classification.
The full process can be summarized as follows:\looseness=-1
\begin{align}
    \label{eq:generative_story} 
    \bm{z} &= \textsf{expl}(\textsf{enc}(\bm{x}; \phi); \gamma),  \\
    \hat{y} &= \textsf{pred}(\bm{x} \odot \bm{z}; \theta), \label{eq:predictor}
\end{align}
where $\phi, \gamma, \theta$ are trainable parameters.
To ensure that the explainer does not select all tokens (i.e., $z_i = 1, \forall i$), sparsity is usually encouraged in the rationale extraction. Moreover, explainers can also be encouraged to
select contiguous words, as there is some evidence that it improves readibility~\citep{jain-etal-2020-learning}. These desired properties may be encouraged via regularization terms during training~\citep{lei-etal-2016-rationalizing, bastings-etal-2019-interpretable}, or via application of sparse mappings~\citep{treviso-martins-2020-explanation, guerreiro-martins-2021-spectra}.

In this work, we will focus specifically on the SPECTRA rationalizer~\citep{guerreiro-martins-2021-spectra}: this model leverages an explainer that extracts a deterministic structured mask $\bm{z}$ by solving a constrained inference problem with SparseMAP~\citep{niculae2018sparsemap}. SPECTRA has been shown to achieve comparable performance with other rationalization approaches, in terms of end-task performance, plausibility with human explanations, and robustness to input perturbation~\citep{chen-etal-2022-rationalization}. Moreover, it is easier to train than other stochastic alternatives~\citep{lei-etal-2016-rationalizing, bastings-etal-2019-interpretable}, and, importantly, it allows for simple control over the properties of the rationales, such as sparsity via its constrained inference formulation: by setting a budget $B$ on the rationale extraction, SPECTRA ensures that the rationale size will not exceed  $\lceil B n \rceil$ tokens.

\subsection{Counterfactuals}

In NLP, counterfactuals refer to alternative texts that describe a different outcome than what is encoded in a given factual text. 
Prior works~\citep{verma2020counterfactual} have focused on developing methods for generating counterfactuals that adhere to 
several key
properties, 
including:
\begin{itemize}
    \item \textbf{Validity}: 
    the generated counterfactuals should encode a different label from the original text.\looseness=-1

    \item \textbf{Closeness}: 
    the changes made to the text should be small, not involving large-scale rewriting of the input. 
    
    \item \textbf{Fluency}: 
    the generated counterfactuals should be coherent and grammatically correct.
    
    \item \textbf{Diversity}: 
    the method should generate a wide range of counterfactuals with diverse characteristics, rather than only a limited set of variations.
\end{itemize}

While many methods for automatic counterfactual generation exist~\citep{wu-etal-2021-polyjuice,robeer-etal-2021-generating-realistic,dixit2022core}, 
our work is mostly related to MiCE~\citep{ross-etal-2021-explaining}, which generates counterfactuals in a two stage process that involves masking the top-$k$ tokens with the highest $\ell_1$ gradient attribution of a pre-trained classifier, and infilling tokens for masked position with a T5-based model~\citep{raffel2020exploring}. 
MiCE further refines the resultant counterfactual with a binary search procedure to seek strictly \emph{minimal} edits. However, this process is computationally expensive and, as we show in \S\ref{subsec:results_counterfactual_generation}, directly optimizing for closeness can lead to counterfactuals that are less valid, fluent, and diverse. Next, we present an alternative method that overcomes these limitations while still producing counterfactuals that are close to original inputs.\looseness=-1

\section{\crestgen}
\label{sec:crest_gen}

We now introduce CREST (ContRastive Edits with Sparse raTionalization), a framework that combines selective rationalization and counterfactual text generation. CREST has two key components: 
(i) \textbf{\crestgen} offers a controlled approach to generating counterfactuals, which we show are valid, fluent, and diverse (\S\ref{subsec:results_counterfactual_generation}); 
and (ii)~\textbf{\crestrat} leverages these counterfactuals through a novel regularization technique encouraging agreement between rationales for original and counterfactual examples. 
We demonstrate that combining these two components leads to models that are more robust 
(\S\ref{subsec:results_training_with_counterfactuals}) and interpretable
(\S\ref{subsec:interpretability_analysis}).
We describe \crestgen below and \crestrat in \S\ref{sec:crest_rationalization}.

Formally, let $\bm{x} = \langle x_1, ..., x_n\rangle$ represent a factual input text with a label $y_f$. 
We define a counterfactual as an input $\tilde{\bm{x}} = \langle x_1, ..., x_m \rangle$ labeled with $y_c$ such that $y_f \neq y_c$. 
A counterfactual generator is a mapping that transforms the original text $\bm{x}$ to a counterfactual $\tilde{\bm{x}}$. 
Like MiCE, our approach for generating counterfactuals consists of two stages, as depicted in Figure~\ref{fig:intro_fig}: the mask and the edit stages.

\begin{table*}[t]
    \small
    \centering
    \begin{tabular}{l c@{\tquad}c@{\tquad}c@{\tquad}c@{\tquad}c c@{\ \ } c@{\tquad}c@{\tquad}c@{\tquad}c@{\tquad}c}
        \toprule
        
        & \multicolumn{5}{c}{\bf IMDB} & & \multicolumn{5}{c}{\bf SNLI} \\
        \cmidrule{2-6} 
        \cmidrule{8-12}
        \bf Method & \bf val. $\uparrow$ & \bf fl. $\downarrow$  & \bf div. $\downarrow$ & \bf clo. $\downarrow$ & \bf \#tks & & 
                     \bf val. $\uparrow$ & \bf fl. $\downarrow$  & \bf div. $\downarrow$ & \bf clo. $\downarrow$ & \bf \#tks \\
        \midrule
        Chance  baseline        & 50.20 & -     & - & - & -             & & 52.70 & - & - & - & -  \\
        
        References          & 97.95 & 66.51 & -     & -    & 184.4      & & 96.75 & 63.52 & - & - & 7.5  \\
        
        Manual edits        & 93.44 & 72.89 & 81.67 & 0.14 & 183.7      & & 93.88 & 65.25 & 35.82 & 0.42 & 7.7 \\
        \cdashlinelr{1-12}
        
        PWWS                                & 28.07 & 101.91 & 74.56 & \bf 0.16 & 179.0     & & 17.97 & 160.11 & 31.81 & 0.36 & 6.8 \\
        
        CFGAN                               & -     & -  & - & - & -                    	& & 34.46 & 155.84 & 68.94 & \bf 0.23 & 7.0 \\
        
        PolyJuice                           & 36.69 & 68.59 & 56.41 & 0.45 & 94.6           & & 41.80 & 62.62 & 39.01 & 0.40 & 11.6\\

        MiCE (bin. search)          & 72.13 & 76.72 & 73.76 & 0.20 & 171.3      & & 76.17 & 63.94 & 42.18 & 0.35 & 7.9 \\
        
        \cdashlinelr{1-12}
        
        MiCE (30\% mask)              & 76.80 & 79.35 & 49.64 & 0.39 & 161.3        & & 77.26 & \bf 59.71 & 34.08 & 0.40 & 8.3 \\
        
        MiCE (50\% mask)              & 83.20 & 89.92 & \bf 20.71 & 0.65 & 115.7    & & \bf 84.48 & 68.32 & \bf 24.27 & 0.52 & 7.6 \\

        \cdashlinelr{1-12}
        
        CREST (30\% mask)                  & 75.82 & 67.29 & 57.58 & 0.33 & 180.9           & & 75.45 & 62.00 & 41.36 & 0.29 & 7.4 \\
        CREST (50\% mask)                  & \bf 93.24 & \bf 50.69 & 23.08 & 0.66 & 193.9   & & 81.23 & 62.60 & 30.53 & 0.41 & 7.3 \\
        \bottomrule
    \end{tabular}
    \caption{
    Intrinsic evaluation of counterfactuals generated by various methods. 
    Validity is computed as the accuracy of an off-the-shelf RoBERTa-base classifier in relation to the gold counterfactual label (not available for PWWS and PolyJuice); 
    fluency is determined by the perplexity score given by GPT-2 large; diversity is computed with self-BLEU; and closeness is reported by the (normalized) edit distance to the factual input.
    In addition, we report the average number of tokens in the input. \looseness=-1
    }
    \label{tab:results_counterfactuals_methods}
\end{table*}

\paragraph{Mask stage.} We aim to find a mask vector $\bm{z} \in \{0,1\}^n$ such that tokens $x_i$ associated with $z_i = 1$ are relevant for the factual prediction $\hat{y}_f$ of a particular classifier $C$. To this end, we employ a SPECTRA rationalizer as the \textbf{masker}. Concretely, we pre-train a SPECTRA rationalizer on the task at hand with a budget constraint $B$, and define the mask as the rationale vector $\bm{z} \in \{0, 1\}^n$ (see \S\ref{subsec:vanilla_rationalizer}).

\paragraph{Edit stage.} Here, we create edits by infilling the masked positions using an \textbf{editor} module $G$, such as a masked language model: $\tilde{\bm{x}} \sim G_{\mathrm{LM}}(\bm{x} \odot \bm{z})$. In order to infill spans rather than single tokens, we follow MiCE and use a T5-based model to infill spans for masked positions.
During training, we fine-tune the editor to infill original spans of text by prepending gold target labels $y_f$ to original inputs.
In order to generate counterfactual edits at test time, we prepend a counterfactual label $y_c$ instead, and sample counterfactuals using beam search.

Overall, our procedure differs from that of MiCE in the mask stage: instead of extracting a mask via gradient-based attributions and subsequent binary search, we leverage SPECTRA to find an optimal mask. Interestingly, by doing so, we not only avoid the computationally expensive binary search procedure, but we also open up new opportunities: as our masking process is differentiable, we can optimize our masker to enhance the quality of both the counterfactuals (\S\ref{subsec:results_counterfactual_generation}) and the selected rationales (\S\ref{subsec:interpretability_analysis}). 
We will demonstrate the latter with our proposed \crestrat setup~(\S\ref{sec:crest_rationalization}). All implementation details for the masker and the editor can be found in \S\ref{sec:CREST_details}. \looseness=-1

\section{Evaluating CREST Counterfactuals}
\label{sec:counterfactual_generation}

This section presents an extensive comparison of counterfactuals generated by different methods.

\begin{figure*}[t]
    \centering
    \begin{tikzpicture}[scale=1]
    \begin{axis}[
        cycle list name=my style,
        xlabel={Budget percentage},
        ylabel={Validity (\%)},
        ymin=40, ymax=96,
        enlarge x limits=0.05,
        enlarge y limits=0.05,
        legend cell align={left},
        xtick={0.1, 0.2, 0.3, 0.4, 0.5},
        ytick={40, 54, 68, 82, 96},
        font=\small,
        tick label style={font=\scriptsize},
        legend pos=south east,
        legend style={font=\scriptsize},
        grid=both,
        grid style={line width=.1pt, draw=gray!15, dashed},
        every axis plot/.append style={very thick},
        width=0.32\linewidth,
        height=0.25\linewidth,
    ]
        \addplot
        coordinates {
            (0.1, 42.01) (0.2, 60.25) (0.3, 77.05) (0.4, 83.20) (0.5, 93.24)
        };
        \addplot
        coordinates {
            (0.1, 50.00) (0.2, 70.70) (0.3, 76.43) (0.4, 85.04) (0.5, 93.85)
        };
        \addplot
        coordinates {
            (0.1, 60.45) (0.2, 71.52) (0.3, 86.07) (0.4, 88.93) (0.5, 94.47)
        }; 
    \end{axis}
\end{tikzpicture}
    \quad
    \begin{tikzpicture}[scale=1]
    \begin{axis}[
        cycle list name=my style,
        xlabel={Budget percentage},
        ylabel={Fluency (ppl.)},
        ymin=50, ymax=82,
        enlarge x limits=0.05,
        enlarge y limits=0.05,
        legend cell align={left},
        xtick={0.1, 0.2, 0.3, 0.4, 0.5},
        ytick={50, 58, 66, 74, 82},
        grid=both,
        grid style={line width=.1pt, draw=gray!15, dashed},
        every axis plot/.append style={very thick},
        tick label style={font=\scriptsize},
        width=0.32\linewidth,
        height=0.25\linewidth,
        font=\small,
    ]
        \addplot
        coordinates {
            (0.1, 75.38) (0.2, 74.33) (0.3, 70.25) (0.4, 61.55) (0.5, 52.86)
        };
        \addplot
        coordinates {
            (0.1, 75.41) (0.2, 74.50) (0.3, 70.06) (0.4, 61.40) (0.5, 52.74)
        };
        \addplot
        coordinates {
            (0.1, 77.79) (0.2, 77.67) (0.3, 71.63) (0.4, 62.29) (0.5, 52.89)
        }; 
    \end{axis}
\end{tikzpicture}
    \quad
    \begin{tikzpicture}[scale=1]
    \begin{axis}[
        cycle list name=my style,
        xlabel={Budget percentage},
        ylabel={Closeness (\%)},
        ymin=10, ymax=70,
        enlarge x limits=0.05,
        enlarge y limits=0.05,
        xtick={0.1, 0.2, 0.3, 0.4, 0.5},
        ytick={10, 25, 40, 55, 70},
        legend cell align={left},
        legend pos=north west,
        legend style={font=\scriptsize, draw=none, fill=none, style={row sep=-0.1pt}},
        legend image post style={scale=0.5},
        grid=both,
        grid style={line width=.1pt, draw=gray!15, dashed},
        every axis plot/.append style={very thick},
        tick label style={font=\scriptsize},
        width=0.32\linewidth,
        height=0.25\linewidth,
        font=\small,
    ]
        \addplot
        coordinates {
            (0.1, 11) (0.2, 23) (0.3, 35) (0.4, 46) (0.5, 68)
        };
        \addplot
        coordinates {
            (0.1, 10) (0.2, 23) (0.3, 35) (0.4, 45) (0.5, 67)
        };
        \addplot
        coordinates {
            (0.1, 11) (0.2, 22) (0.3, 36) (0.4, 46) (0.5, 67)
        }; 
        \legend{original, augmented, finetuned}
    \end{axis}
\end{tikzpicture}
    \caption{Sparsity analysis of \crestgen on IMDB with different budget percentages.
    The \textit{original} curves show the performance of CREST without any changes, while the \textit{augmented} and \textit{finetuned} curves show the performance of CREST when using manually crafted counterfactuals for data augmentation or finetuning, respectively.\looseness=-1
    }
    \label{fig:mask_analysis}
\end{figure*}
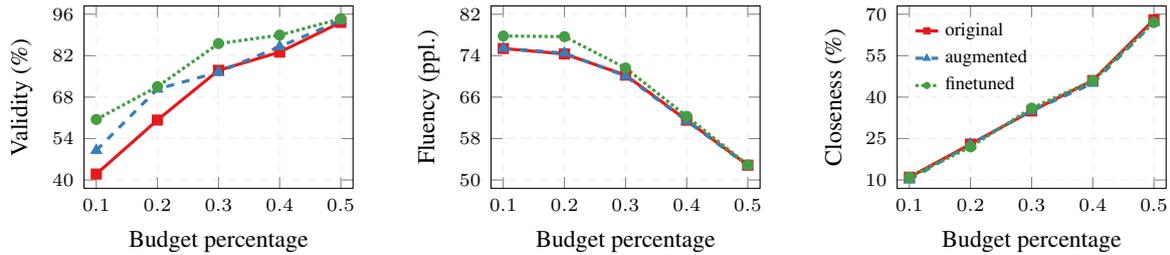

\subsection{Experimental Setting}
\paragraph{Data and evaluation.} 
We experiment with our counterfactual generation framework on two different tasks: 
sentiment classification using IMDB~\citep{maas-etal-2011-learning} and natural language inference (NLI) using SNLI~\citep{bowman-etal-2015-large}. 
In sentiment classification, we only have a single input to consider, while NLI inputs consist of a premise and a hypothesis, which we concatenate to form a single input.
To assess the quality of our automatic counterfactuals, 
we compare them to manually crafted counterfactuals in the revised IMDB and SNLI datasets created by \citet{Kaushik2020Learning}. 
More dataset details can be found in \S\ref{sec:datasets_stats}.

\paragraph{Training.} We employ a SPECTRA rationalizer with a T5-small architecture as the masker, and train it for 10 epochs on the full IMDB and SNLI datasets. 
We also use a T5-small architecture for the editor, and train it for 20 epochs with early stopping, following the same training recipe as MiCE. Full training details can be found in \S\ref{subsec:CREST_details_SPECTRA_training}.

\paragraph{Generation.} As illustrated in Figure~\ref{fig:intro_fig}, at test time we generate counterfactuals by prepending a contrastive label to the input and passing it to the editor. For sentiment classification, this means switching between positive and negative labels. 
For NLI, in alignment with \citet{dixit2022core}, we adopt a refined approach by restricting the generation of counterfactuals to entailments and contradictions only, therefore ignoring neutral examples, which have a subtle semantic meaning. In contrast, our predictors were trained using neutral examples, and in cases where they predict the neutral class, we default to the second-most probable class.

\paragraph{Baselines.} 
We compare our approach with four open-source baselines that generate counterfactuals: PWWS~\citep{ren-etal-2019-generating}, PolyJuice~\citep{wu-etal-2021-polyjuice},
CounterfactualGAN~\citep{robeer-etal-2021-generating-realistic},\footnote{Despite many attempts, CounterfactualGAN did not converge on IMDB, possibly due to the long length of the inputs.} and MiCE~\citep{ross-etal-2021-explaining}. In particular, to ensure a fair comparison with MiCE, we apply three modifications to the original formulation: 
(i)~we replace its RoBERTa classifier with a T5-based classifier (as used in SPECTRA); 
(ii)~we disable its validity filtering;\footnote{MiCE with binary search uses implicit validity filtering throughout the search process to set the masking percentage.}
(iii)~we report results with and without the binary search procedure by fixing the percentage of masked tokens.

\paragraph{Metrics.} 
To determine the general \textbf{validity} of counterfactuals, we report the accuracy of an off-the-shelf RoBERTa-base classifier available in the HuggingFace Hub.\footnote{\texttt{mtreviso/roberta-base-imdb}, \texttt{mtreviso/roberta-base-snli}.}
Moreover, we measure \textbf{fluency} using perplexity scores from GPT-2 large~\citep{radford2019language} and \textbf{diversity} with self-BLEU~\citep{zhu2018texygen}. Finally, we quantify the notion of \textbf{closeness} by computing the normalized edit distance to the factual input and the average number of tokens in the document.

\subsection{Results} \label{subsec:results_counterfactual_generation}

Results are presented in Table~\ref{tab:results_counterfactuals_methods}.
As expected, manually crafted counterfactuals achieve high validity, significantly surpassing the chance baseline and establishing a reliable reference point.
For IMDB, we find that CREST outperforms other methods by a wide margin in terms of validity and fluency. 
At the same time, CREST's validity is comparable to the manually crafted counterfactuals, while surprisingly deemed more fluent by GPT-2. 
Moreover, we note that our modification of disabling MiCE's minimality search leads to counterfactuals that are more valid and diverse but less fluent and less close to the original inputs. 

For SNLI, this modification allows MiCE to achieve the best overall scores, closely followed by CREST. 
However, when controlling for closeness, we observe that CREST outperforms MiCE: at closeness of $\sim$0.30, CREST (30\% mask) outperforms MiCE with binary search in terms of fluency and diversity. 
Similarly, at a closeness of $\sim$0.40, CREST (50\% mask) surpasses MiCE (30\% mask) across the board. 
As detailed in \S\ref{sec:validity_x_closeness}, CREST's counterfactuals are more valid than MiCE's for all closeness bins lower than 38\%. 
We provide examples of counterfactuals produced by CREST and MiCE in Appendix~\ref{app:examples_counterfactuals}.
Finally, we note that CREST is highly affected by the masking budget, which we explore further next.

\paragraph{Sparsity analysis.} 
We investigate how the number of edits affects counterfactual quality by training maskers with increasing budget constraints (as described in \S\ref{subsec:vanilla_rationalizer}). 
The results in Figure~\ref{fig:mask_analysis} show that with increasing masking percentage, generated counterfactuals become less textually similar to original inputs (i.e., less close) but more valid and fluent. This inverse relationship demonstrates that strict minimality, optimized for in methods like MiCE, comes with tradeoffs in counterfactual quality, and that the sparsity budget in \crest can be used to modulate the trade-off between validity and closeness.
In Figure~\ref{fig:mask_analysis} we also examine the benefit of manually crafted counterfactuals in two ways: (i) using these examples as additional training data; and (ii) upon having a trained editor, further fine-tuning it with these manual counterfactuals.
The results suggest that at lower budget percentages, exploiting a few manually crafted counterfactuals to fine-tune CREST can improve the validity of counterfactuals without harming fluency.

\paragraph{Validity filtering.} 
As previously demonstrated by \citet{wu-etal-2021-polyjuice} and \citet{ross-etal-2022-tailor}, it is possible to filter out potentially disfluent or invalid counterfactuals by passing all examples to a classifier and discarding the subset with incorrect predictions. 
In our case, we use the predictor associated with the masker as the classifier. 
We found find that applying this filtering increases the validity of IMDB counterfactuals from 75.82 to 86.36 with $B=0.3$, and from 93.24 to 97.36 with $B=0.5$. For SNLI, validity jumps from 75.45 to 96.39 with $B=0.3$, and from 81.23 to 96.67 with $B=0.5$.
These results indicate that CREST can rely on its predictor to filter out invalid counterfactuals, a useful characteristic for doing data augmentation, as we will see in \S\ref{subsec:results_training_with_counterfactuals}.\looseness=-1

\begin{figure}[t]
    \centering
    \includegraphics[width=0.85\columnwidth]{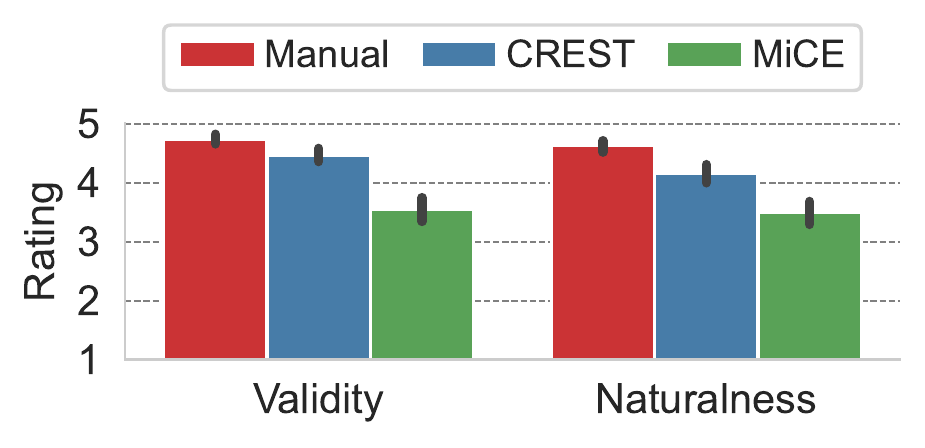}
    \vspace*{-0.2cm}
    \caption{
    Human study results for counterfactuals produced manually and automatically (CREST and MiCE).
    }
    \label{fig:human_study_results}
\end{figure}

\subsection{Human Study}

We conduct a small-scale human study to evaluate the quality of counterfactuals produced by MiCE and CREST with 50\% masking percentage. Annotators were tasked with rating counterfactuals' \emph{validity} and \emph{naturalness} (e.g., based on style, tone, and grammar), each using a 5-point Likert scale. Two fluent English annotators rated 50 examples from the IMDB dataset, and two others rated 50 examples from SNLI. We also evaluate manually created counterfactuals to establish a reliable baseline. More annotation details can be found in \S\ref{sec:human_study}.

The study results, depicted in Figure~\ref{fig:human_study_results}, show that humans find manual counterfactuals to be more valid and natural compared to automatically generated ones. Furthermore, CREST's counterfactuals receive higher ratings for validity and naturalness compared to MiCE, aligning with the results obtained from automatic metrics.

\section{\crestrat}
\label{sec:crest_rationalization}

Now that we have a method that generates high-quality counterfactual examples, a natural step is to use these examples for data augmentation.
However, vanilla data augmentation does not take advantage of the paired structure of original/contrastive examples and instead just treats them as individual datapoints. 
In this section, we present CREST's second component, \crestrat (illustrated in Figure~\ref{fig:intro_fig_2}), which leverages the relationships between factual and counterfactual inputs through a SPECTRA rationalizer with an \textbf{agreement regularization} strategy, described next.

\subsection{Linking Counterfactuals and Rationales} \label{subsec:linking_counterfactuals_and_rationales}
We propose to incorporate counterfactuals into a model's functionality by taking advantage of the fully differentiable rationalization setup.
Concretely, we decompose a rationalizer into two flows, as depicted in Figure~\ref{fig:intro_fig_2}: a \textbf{factual flow} that receives factual inputs $\bm{x}$ and outputs a factual prediction $\hat{y}$, and a \textbf{counterfactual flow} that receives counterfactual inputs $\tilde{\bm{x}}$ and should output a counterfactual prediction $\tilde{y} \ne \hat{y}$. 
As a by-product of using a rationalizer, we also obtain a factual rationale $\bm{z} \in \{0, 1\}^n$ for $\bm{x}$ and a counterfactual rationale $\tilde{\bm{z}} \in \{0, 1\}^m$ for $\tilde{\bm{x}}$, where $n = |\bm{x}|$ and $m = |\tilde{\bm{x}}|$.

\paragraph{Training.} Let $\Theta = \{\phi, \gamma, \theta\}$ represent the trainable parameters of a rationalizer (defined in \S\ref{subsec:vanilla_rationalizer}). 
We propose the following loss function:
\begin{align} \label{eq:training_crest_rationalization}
    \mathcal{L}(\Theta) &= \mathcal{L}_f(y_f, \hat{y}(\Theta)) + \alpha \mathcal{L}_c(y_c, \tilde{y}(\Theta)) \\
    &+ \lambda \Omega(\bm{z}(\Theta), \tilde{\bm{z}}(\Theta)), \nonumber
\end{align}
where $\mathcal{L}_f(\cdot)$ and $\mathcal{L}_c(\cdot)$ represent cross-entropy losses for the factual and counterfactual flows, respectively, and $\Omega(\cdot)$ is a novel penalty term to encourage factual and counterfactual rationales to focus on the same positions, as defined next. $\alpha \in \mathbb{R}$ and $\lambda \in \mathbb{R}$ are hyperparameters.

\begin{table*}[t]
    \small
    \centering
    \begin{tabular}{lccccccc}
        \toprule
        \bf Setup & \bf \cellcolor{cid}{IMDB} & \bf \cellcolor{ccon}{rIMDB} & \bf \cellcolor{ccon}{cIMDB} & \bf \cellcolor{cood}{RotTom} & \bf \cellcolor{cood}{SST-2} & \bf \cellcolor{cood}{Amazon} & \bf \cellcolor{cood}{Yelp}\\

        \midrule
        $F$                    & \underline{91.1} \scriptsize{$\pm$ 0.3} & 91.4 \scriptsize{$\pm$ 0.8} & 88.5 \scriptsize{$\pm$ 0.9} & 76.5 \scriptsize{$\pm$ 1.6}  & 79.8 \scriptsize{$\pm$ 1.6}  & 86.0 \scriptsize{$\pm$ 0.7}  & 88.5 \scriptsize{$\pm$ 0.7} \\

        \cdashlinelr{1-8}
        \multicolumn{8}{l}{\textit{With data augmentation:}} \\

        $F + C_H$       & 90.9 \scriptsize{$\pm$ 0.5} & \textbf{92.9} \scriptsize{$\pm$ 0.9} & \textbf{90.4} \scriptsize{$\pm$ 1.6} & 76.6 \scriptsize{$\pm$ 1.5}  & \underline{80.7} \scriptsize{$\pm$ 1.3}  & 86.3 \scriptsize{$\pm$ 1.0}  & \underline{89.1} \scriptsize{$\pm$ 1.2} \\

        $F + C_{S,V}$   & 91.0 \scriptsize{$\pm$ 0.2} & 91.2 \scriptsize{$\pm$ 1.0} & 89.3 \scriptsize{$\pm$ 0.8} & \underline{76.8} \scriptsize{$\pm$ 0.9} & 79.3 \scriptsize{$\pm$ 0.3} & 85.2 \scriptsize{$\pm$ 0.9} & 88.0 \scriptsize{$\pm$ 1.0}  \\
        $F + C_{S}$     & 90.8 \scriptsize{$\pm$ 0.2} & 91.6 \scriptsize{$\pm$ 1.3} & 89.2 \scriptsize{$\pm$ 0.4} & 76.7 \scriptsize{$\pm$ 1.0} & 80.6 \scriptsize{$\pm$ 0.6} & \underline{86.4} \scriptsize{$\pm$ 0.6} & \underline{89.1} \scriptsize{$\pm$ 0.5} \\
        
        \cdashlinelr{1-8}
        \multicolumn{8}{l}{\textit{With agreement regularization:}} \\
        
        $F\ \&\ C_{S, V}$      & 90.7 \scriptsize{$\pm$ 0.5} & \underline{92.2} \scriptsize{$\pm$ 0.7} & 88.9 \scriptsize{$\pm$ 1.0} & 76.3 \scriptsize{$\pm$ 1.4} & 80.2 \scriptsize{$\pm$ 1.3} & 86.3 \scriptsize{$\pm$ 0.7} & 88.9 \scriptsize{$\pm$ 0.7}  \\
        $F\ \&\ C_{S}$       & \textbf{91.2} \scriptsize{$\pm$ 0.5} & \textbf{92.9} \scriptsize{$\pm$ 0.5} & \underline{89.7} \scriptsize{$\pm$ 1.1} & \textbf{77.3} \scriptsize{$\pm$ 2.3} & \textbf{81.1} \scriptsize{$\pm$ 2.4} & \textbf{86.8} \scriptsize{$\pm$ 0.8} & \textbf{89.3} \scriptsize{$\pm$ 0.7}  \\
        
        \bottomrule
    \end{tabular}
    \caption{
    Accuracy of SPECTRA trained on IMDB and evaluated on \colorbox{cid}{in-domain}, \colorbox{ccon}{contrast}, and \colorbox{cood}{out-of-domain} datasets. We present mean and std. values across five random seeds. Values in \textbf{bold}: top results; \underline{underlined}: second-best.
    } 
    \label{tab:results_rationalizers_in_domain}
\end{table*}

\paragraph{Agreement regularization.} 
To produce paired rationales for both the factual and counterfactual flows, we incorporate regularization terms into the training of a rationalizer to encourage the factual explainer to produce rationales similar to those originally generated by the \emph{masker} $\bm{z}^\star$, and the counterfactual explainer to produce rationales that focus on the tokens modified by the \emph{editor} $\tilde{\bm{z}}^\star$. We derive the ground truth counterfactual rationale $\tilde{\bm{z}}^\star$ by aligning $\bm{x}$ to $\tilde{\bm{x}}$ and marking tokens that were inserted or substituted as $1$, and others as $0$. The regularization terms are defined as:
\begin{align}
    \Omega(\bm{z}, \tilde{\bm{z}}) &= \norm{\bm{z}(\Theta) - \bm{z}^\star}_2^2 + \norm{\tilde{\bm{z}}(\Theta) - \tilde{\bm{z}}^\star}_2^2. \label{eq:explainer_reg}
\end{align}
 
To allow the counterfactual rationale $\tilde{\bm{z}}$ to focus on all important positions in the input, we adjust the budget for the counterfactual flow based on the length of the synthetic example produced by the counterfactual generator. Specifically, we multiply the budget by a factor of $\frac{\norm{\tilde{\bm{z}}^\star}_0}{\norm{\bm{z}^\star}_0}$.

\section{Exploiting Counterfactuals for Training}
In this section, we evaluate the effects of incorporating \crest-generated counterfactuals into training by comparing a vanilla data augmentation approach with our \crestrat approach. 
We compare how each affects model robustness~(\S\ref{subsec:results_training_with_counterfactuals}) and interpretability (\S\ref{subsec:interpretability_analysis}).

\subsection{Experimental Setting}
We use the IMDB and SNLI datasets to train SPECTRA rationalizers with and without counterfactual examples, and further evaluate on \colorbox{cid}{in-domain}, \colorbox{ccon}{contrast} and \colorbox{cood}{out-of-domain} (OOD) datasets. 
For IMDB, we evaluate on the \colorbox{ccon}{revised IMDB}, \colorbox{ccon}{contrast IMDB}, \colorbox{cood}{RottenTomatoes}, \colorbox{cood}{SST-2}, \colorbox{cood}{Amazon Polarity}, and \colorbox{cood}{Yelp}.
For SNLI, we evaluate on the \colorbox{cid}{Hard SNLI}, \colorbox{ccon}{revised SNLI}, \colorbox{ccon}{break}, \colorbox{cood}{MultiNLI}, and \colorbox{cood}{Adversarial NLI}. 
Dataset details can be found in \S\ref{sec:datasets_stats}.
To produce \crest counterfactuals, which we refer to as ``synthetic'', we use a 30\% masking budget as it provides a good balance between validity, fluency, and closeness (\textit{cf.} Figure~\ref{fig:mask_analysis}). 
We tune the counterfactual loss ($\alpha$) and agreement regularization ($\lambda$) weights on the dev set. We report results with $\alpha = 0.01$ and $\lambda = 0.001$ for IMDB, and $\alpha = 0.01$ and $\lambda = 0.1$ for SNLI.

\begin{table*}[t]
    \small
    \centering
    \begin{tabular}{lccccccc}
        \toprule
        \textbf{Setup} & \bf \cellcolor{cid}{SNLI} & \bf \cellcolor{cid}{SNLI-h} & \bf \cellcolor{ccon}{rSNLI} & \bf \cellcolor{ccon}{break} & \bf \cellcolor{cood}{MNLI-m} & \bf \cellcolor{cood}{MNLI-mm} & \bf \cellcolor{cood}{ANLI} \\

        \midrule
        
        $F$               & 86.6 \scriptsize{$\pm$ 0.2} & 73.7 \scriptsize{$\pm$ 0.2} & 71.1 \scriptsize{$\pm$ 0.8} & 69.5 \scriptsize{$\pm$ 1.5} & \textbf{64.6} \scriptsize{$\pm$ 1.1} & 65.9 \scriptsize{$\pm$ 0.9} & \underline{32.6} \scriptsize{$\pm$ 0.7} \\

        \cdashlinelr{1-8}
        \multicolumn{8}{l}{\textit{With data augmentation:}} \\
        
        $F + C_H$        & 86.6 \scriptsize{$\pm$ 0.3} & 74.9 \scriptsize{$\pm$ 1.1} & \textbf{72.4} \scriptsize{$\pm$ 0.3} & \underline{70.1} \scriptsize{$\pm$ 1.9} & 64.2 \scriptsize{$\pm$ 0.9} & 65.8 \scriptsize{$\pm$ 0.9} & 31.8 \scriptsize{$\pm$ 0.4} \\

        $F + C_{S,V}$    & 86.5 \scriptsize{$\pm$ 0.3} & \textbf{75.8} \scriptsize{$\pm$ 1.2} & \underline{71.8} \scriptsize{$\pm$ 1.0} & 69.1 \scriptsize{$\pm$ 2.0} & 64.4 \scriptsize{$\pm$ 0.3} & 65.9 \scriptsize{$\pm$ 0.4} & 32.2 \scriptsize{$\pm$ 0.2} \\
        
        $F + C_{S}$     & 86.6 \scriptsize{$\pm$ 0.3} & 74.7 \scriptsize{$\pm$ 1.1} & 71.6 \scriptsize{$\pm$ 0.8} & \textbf{71.2} \scriptsize{$\pm$ 1.4} & \underline{64.5} \scriptsize{$\pm$ 0.4} & \textbf{66.4} \scriptsize{$\pm$ 0.6} & 32.2 \scriptsize{$\pm$ 1.0} \\

        \cdashlinelr{1-8}
        \multicolumn{3}{l}{\textit{With agreement regularization:}} \\
        $F\ \&\ C_{S, V}$     & \textbf{86.8} \scriptsize{$\pm$ 0.1} & 75.3 \scriptsize{$\pm$ 0.8} & 66.8 \scriptsize{$\pm$ 0.7} & 68.2 \scriptsize{$\pm$ 2.1} & \textbf{64.6} \scriptsize{$\pm$ 0.7} & \underline{66.1} \scriptsize{$\pm$ 0.6} & \textbf{32.8} \scriptsize{$\pm$ 0.6}  \\
        
        $F\ \&\ C_{S}$      & \underline{86.6} \scriptsize{$\pm$ 0.1} & \underline{75.5} \scriptsize{$\pm$ 1.3} & 67.0 \scriptsize{$\pm$ 1.3} & 69.9 \scriptsize{$\pm$ 1.7} & 64.2 \scriptsize{$\pm$ 1.1} & 66.0 \scriptsize{$\pm$ 0.7} & 32.5 \scriptsize{$\pm$ 0.5}\\
        
        \bottomrule
    \end{tabular}
    \caption{Accuracy of SPECTRA trained on SNLI and evaluated on \colorbox{cid}{in-domain}, \colorbox{ccon}{contrast}, and \colorbox{cood}{out-of-domain} datasets. We present mean and std. values across five random seeds. 
    Values in \textbf{bold}: top results; \underline{underlined}: second-best.
    }   
    \label{tab:results_rationalizers_in_domain_snli_hyp}
\end{table*}

\subsection{Robustness Results} \label{subsec:results_training_with_counterfactuals}

Tables~\ref{tab:results_rationalizers_in_domain} and~\ref{tab:results_rationalizers_in_domain_snli_hyp} show results for counterfactual data augmentation and agreement regularization for IMDB and SNLI, respectively.
We compare a standard SPECTRA trained on factual examples ($F$) with other SPECTRA models trained on augmentated data from human-crafted counterfactuals ($F + C_H$) and synthetic counterfactuals generated by CREST ($F + C_S$), which we additionally post-process to drop invalid examples~($F + C_{S,V}$). 

\paragraph{Discussion.} 
As shown in Table~\ref{tab:results_rationalizers_in_domain}, \crestrat ($F\ \&\ C_S$) consistently outperforms vanilla counterfactual augmentation ($F + C_S$) on all sentiment classification datasets. 
It achieves the top results on the full IMDB and on all OOD datasets, while also leading to strong results on contrastive datasets---competitive with manual counterfactuals ($F + C_H$). 
When analyzing the performance of \crestrat trained on a subset of valid examples ($F\ \&\ C_{S,V}$) versus the entire dataset ($F\ \&\ C_{S}$), the models trained on the entire dataset maintain a higher level of performance across all datasets. However, when using counterfactuals for data augmentation, this trend is less pronounced, especially for in-domain and contrastive datasets. 
In \S\ref{sec:cf_data_aug_analysis}, we explore the impact of the number of augmented examples on results and find that, consistent with previous research~ \citep{huang-etal-2020-counterfactually,joshi-he-2022-investigation}, augmenting the training set with a small portion of valid and diverse synthetic counterfactuals leads to more robust models, and can even outweigh the benefits of manual counterfactuals.\looseness=-1

Examining the results for NLI in Table~\ref{tab:results_rationalizers_in_domain_snli_hyp}, we observe that both counterfactual augmentation and agreement regularization interchangeably yield top results across datasets.
Remarkably, in contrast to sentiment classification, we achieve more substantial improvements with agreement regularization models when these are trained on valid counterfactuals, as opposed to the full set. 

Overall, these observations imply that \crestrat is a viable alternative to data augmentation for improving model robustness, especially for learning contrastive behavior for sentiment classification. 
In the next section, we explore the advantages of \crestrat for improving model interpretability.

\subsection{Interpretability Analysis}
\label{subsec:interpretability_analysis}

In our final experiments, we assess the benefits of our proposed regularization method on model interpretability. We evaluate effects on rationale quality along three dimensions: plausibility, forward simulability, and counterfactual simulability.

\paragraph{Plausibility.} 
We use the MovieReviews~\citep{deyoung-etal-2020-eraser} and the e-SNLI~\citep{camburu2018esnli} datasets to study the human-likeness of rationales by matching them with human-labeled explanations and measuring their AUC, which automatically accounts for multiple binarization levels.\footnote{We determine the explanation score for a single word by calculating the average of the scores of its word pieces.}

\paragraph{Forward simulability.} 
Simulability measures how often a human agrees with a given classifier when presented with explanations, and many works propose different variants to compute simulability scores in an automatic way~\citep{doshi2017towards,treviso-martins-2020-explanation,hase-etal-2020-leakage,pruthi-etal-2022-evaluating}.
Here, we adopt the framework proposed by \citet{treviso-martins-2020-explanation}, which views explanations as a message between a classifier and a linear student model, and determines simulability as the fraction of examples for which the communication is successful. In our case, we cast a SPECTRA rationalizer as the classifier, use its rationales as explanations, and train a linear student on factual examples of the IMDB and SNLI datasets. 
High simulability scores indicate more understandable and informative explanations.

\paragraph{Counterfactual simulability.} 
Building on the manual simulability setup proposed by \citet{doshi2017towards}, we introduce a new approach to automatically evaluate explanations that interact with counterfactuals.
Formally, let $C$ be a classifier that when given an input $\bm{x}$ produces a prediction $\hat{y}$ and a rationale $\bm{z}$. Moreover, let $G$ be a pre-trained counterfactual editor, which receives $\bm{x}$ and $\bm{z}$ and produces a counterfactual $\tilde{\bm{x}}$ by infilling spans on positions masked according to $\bm{z}$ (e.g., via masking). We define \emph{counterfactual  simulability} as follows: 
\begin{align}
    \frac{1}{N}\sum_{n=1}^N [[ C(\bm{x}_n) \neq C(G(\bm{x}_n \odot \bm{z}_n)) ]], 
\end{align}
where $[[\cdot]]$ is the Iverson bracket notation. Intuitively, counterfactual  simulability measures the ability of a rationale to change the label 
predicted by the classifier when it receives a contrastive edit with infilled tokens by a counterfactual generator as input. 
Therefore, a high counterfactual simulability indicates that the rationale $\bm{z}$ focuses on the highly contrastive parts of the input.

\paragraph{Results.} 
The results of our analysis are shown in Table~\ref{tab:interpretability_analysis}.
We observe that plausibility can substantially benefit from synthetic \crest-generated counterfactual examples, 
especially for a rationalizer trained with our agreement regularization, which outperforms other approaches by a large margin.
Additionally, leveraging synthetic counterfactuals, either via data augmentation or agreement regularization, leads to a high forward simulability score, though by a smaller margin---within the standard deviation of other approaches.
Finally, when looking at counterfactual simulability, we note that models that leverage CREST counterfactuals consistently lead to better rationales. 
In particular, agreement regularization leads to strong results on both tasks while also producing more plausible rationales, showing the efficacy of \crestrat in learning contrastive behavior.

\begin{table*}[t]
    \small
    \centering
    \begin{tabular}{l ccc c ccc}
        \toprule
        & \multicolumn{3}{c}{\bf Sentiment Classification} & & \multicolumn{3}{c}{\bf Natural Language Inference} \\
        \cmidrule{2-4} 
        \cmidrule{6-8}
        \bf Setup & \bf Plausibility & \bf F. sim. & \bf C. sim. & & \bf Plausibility & \bf F. sim. & \bf C. sim. \\
        \midrule
        $F$                           & 0.6733  \scriptsize{$\pm$ 0.02} & \underline{91.70} \scriptsize{$\pm$ 0.92} & 81.18 \scriptsize{$\pm$ 2.79} & & 0.7735 \scriptsize{$\pm$ 0.00} & 59.26 \scriptsize{$\pm$ 0.41} & 70.01 \scriptsize{$\pm$ 0.44} \\

        \cdashlinelr{1-8}
        \multicolumn{3}{l}{\textit{With data augmentation:}} \\
        $F + C_H$                     & 0.6718  \scriptsize{$\pm$ 0.04} & 91.44 \scriptsize{$\pm$ 1.46} & 80.53 \scriptsize{$\pm$ 4.17} & & 0.7736 \scriptsize{$\pm$ 0.01} & \underline{59.51} \scriptsize{$\pm$ 0.86} & 69.90 \scriptsize{$\pm$ 0.57} \\
        
        $F + C_S$                & \underline{0.6758}  \scriptsize{$\pm$ 0.01} & 91.68 \scriptsize{$\pm$ 0.59} & \underline{84.54} \scriptsize{$\pm$ 1.09} & & \underline{0.7779} \scriptsize{$\pm$ 0.00} & \textbf{59.54} \scriptsize{$\pm$ 0.08} & \textbf{70.76} \scriptsize{$\pm$ 0.54} \\
        
        \cdashlinelr{1-8}
        \multicolumn{3}{l}{\textit{With agreement regularization:}} \\
        
        $F\ \&\ C_S$ & \textbf{0.6904}  \scriptsize{$\pm$ 0.02} & \textbf{91.93} \scriptsize{$\pm$ 0.83} & \textbf{86.43} \scriptsize{$\pm$ 1.56} & & \textbf{0.7808} \scriptsize{$\pm$ 0.00} & 59.31 \scriptsize{$\pm$ 0.20} &  \underline{70.69} \scriptsize{$\pm$ 0.29} \\
        \bottomrule
    \end{tabular}
    \caption{Interpretability analysis of rationalizers trained with CREST-generated counterfactuals, either with data augmentation or agreement regularization. Plausibility represents matching with human rationales, whereas F. sim. and C. sim. represent forward and counterfactual simulability. \textbf{Bold}: top results; \underline{underlined}: second-best. 
    }
    \label{tab:interpretability_analysis}
\end{table*}

\section{Related Works}

\paragraph{Generating counterfactuals.} Existing approaches to generating counterfactuals for NLP use heuristics \citep{ren-etal-2019-generating,ribeiro-etal-2020-beyond}, leverage plug-and-play approaches to controlled generation \citep{madaan2021generate}, or, most relatedly, fine-tune language models to generate counterfactuals \citep{wu-etal-2021-polyjuice,ross-etal-2021-explaining,ross-etal-2022-tailor,robeer-etal-2021-generating-realistic}. For instance,
PolyJuice~\citep{wu-etal-2021-polyjuice} finetunes a GPT-2 model on human-crafted counterfactuals to generate counterfactuals following pre-defined control codes, while CounterfactualGAN~\citep{robeer-etal-2021-generating-realistic} adopts a GAN-like setup. We show that \crestgen outperforms both methods in terms of counterfactual quality. 
Most closely related is MiCE~\citep{ross-etal-2021-explaining}, which also uses a two-stage approach based on a masker and an editor to generate counterfactuals. Unlike MiCE, we propose to relax the minimality constraint and generate masks using selective rationales rather than gradients, resulting not only in higher-quality counterfactuals, but also in a fully-differentiable set-up that allows for further optimization of the masker. 
Other recent work includes Tailor~\citep{ross-etal-2022-tailor}, a semantically-controlled generation system that requires a human-in-the-loop to generate counterfactuals, 
as well as retrieval-based and prompting approaches such as RGF \citep{paranjape-etal-2022-retrieval} and CORE~\citep{dixit2022core}.

\paragraph{Training with counterfactuals.} 
Existing approaches to training with counterfactuals predominantly leverage data augmentation. Priors works have explored how augmenting with both manual~\citep{Kaushik2020Learning,khashabi-etal-2020-bang,huang-etal-2020-counterfactually,joshi-he-2022-investigation} and automatically-generated~\citep{wu-etal-2021-polyjuice,ross-etal-2022-tailor,dixit2022core} counterfactuals affects model robustness. 
Unlike these works, \crestrat introduces a new strategy for training with counterfactuals that leverages the paired structure of original and counterfactual examples, improving model robustness and interpretability compared to data augmentation. Also related is the training objective proposed by \citet{gupta-etal-2021-paired}  to promote consistency across pairs of examples with shared substructures for neural module networks, and the loss term proposed by \citet{teney2020learning} to model the factual-counterfactual paired structured via gradient supervision.
In contrast, \crest can be used to \textit{generate} paired examples, can be applied to non-modular tasks, and does not require second-order derivatives.

\paragraph{Rationalization.} 
There have been many modifications to the rationalization setup to improve task accuracy and rationale quality. Some examples include conditioning the rationalization on pre-specified labels~\citep{yu-etal-2019-rethinking}, using an information-bottleneck formulation to ensure informative rationales~\citep{paranjape-etal-2020-information}, training with human-created rationales \citep{lehman-etal-2019-inferring}, and replacing stochastic variables with deterministic mappings~\citep{guerreiro-martins-2021-spectra}. 
We find that \crestrat, which is fully unsupervised, outperforms standard rationalizers in terms of model robustness and quality of rationales.\looseness=-1

\section{Conclusions}

We proposed CREST, a joint framework for selective rationalization and counterfactual text generation that is capable of producing valid, fluent, and diverse counterfactuals, while being flexible for controlling the amount of perturbations.
We have shown that counterfactuals can be successfully incorporated into a rationalizer, either via counterfactual data augmentation or agreement regularization, to improve model robustness and rationale quality.
Our results demonstrate that CREST successfully bridges the gap between selective rationales and counterfactual examples, addressing the limitations of existing methods and providing a more comprehensive view of a model's predictions.

\section*{Limitations}
Our work shows that CREST is a suitable framework for generating high-quality counterfactuals and producing plausible rationales, and we hope that CREST motivates new research to develop more robust and interpretable models. 
We note, however, two main limitations in our framework. 
First, our counterfactuals are the result of a large language model (T5), and as such, they may carry all the limitations within these models. Therefore, caution should be exercised when making statements about the quality of counterfactuals beyond the metrics reported in this paper, especially if these statements might have societal impacts. 
Second, CREST relies on a rationalizer to produce highlights-based explanations, and therefore it is limited in its ability to answer interpretability questions that go beyond the tokens of the factual or counterfactual input.

\section*{Acknowledgments}
This work was supported by the European Research Council (ERC StG DeepSPIN 758969), by EU's Horizon Europe Research and Innovation Actions (UTTER, contract 101070631), by P2020 project MAIA (LISBOA-01-0247- FEDER045909), by the Portuguese Recovery and Resilience Plan  through project C645008882-00000055 (NextGenAI, Center for Responsible AI), and by contract UIDB/50008/2020. 
We are grateful to Duarte Alves, Haau-Sing Lee, Taisiya Glushkova, and Henrico Brum for the participation in human evaluation experiments.

\bibliography{anthology,custom}
\bibliographystyle{acl_natbib}

\appendix

\section{Datasets} \label{sec:datasets_stats}

The revised IMDB and SNLI datasets, which we refer to as rIMDB and rSNLI respectively, were created by \citet{Kaushik2020Learning}. 
They contain counterfactuals consisting of revised versions made by humans on the Amazon's Mechanical Turk crowdsourcing platform. For both datasets, the authors ensure that (a) the counterfactuals are valid; (b) the edited reviews are coherent; and (c) the counterfactuals do not contain unnecessary modifications. For SNLI, counterfactuals were created either by revising the premise or the hypothesis. We refer to \citep{Kaushik2020Learning} for more details on the data generation process. Table~\ref{tab:datasets_stats} presents statistics for the datasets used for training models in this work.

\begin{table}[!htb]
    \centering
    \small
    \setlength{\tabcolsep}{4pt}
    \begin{tabular}{lrrc@{\ }rrc@{\ }rr}
         \toprule
         & \multicolumn{2}{c}{\bf Train} & & \multicolumn{2}{c}{\bf Val.} & & \multicolumn{2}{c}{\bf Test} \\ 
         \cmidrule{2-3} 
         \cmidrule{5-6} 
         \cmidrule{8-9} 
         \bf Dataset & \bf docs & \bf tks & & \bf docs & \bf tks & & \bf docs & \bf tks \\
         \midrule
         IMDB    & 22.5K   & 6M   & & 2.5K   & 679K & & 25K    & 6M \\
         rIMDB   & 3414    & 629K & & 490    & 92K & & 976     & 180K \\
         \cdashlinelr{1-9}
         SNLI    & 549K    & 12M  & & 10K    & 232K & & 10K    & 231K \\
         rSNLI   & 4165    & 188K & & 500    & 24K & & 1000    & 48K \\
         \bottomrule
    \end{tabular}
    \caption{Datasets statistics.}
    \label{tab:datasets_stats}
\end{table}

Additionally, we incorporate various contrastive and out-of-domain datasets to evaluate our models.
For IMDB, we use the contrast IMDB~\citep{gardner-etal-2020-evaluating}, RottenTomatoes~\citep{pang-lee-2005-seeing}, SST-2~\citep{socher-etal-2013-recursive}, Amazon Polarity and Yelp~\citep{zhang2015character}. 
For SNLI, we evaluate on the Hard SNLI~\citep{gururangan-etal-2018-annotation}, break~\citep{glockner-etal-2018-breaking}, MultiNLI~\citep{williams-etal-2018-broad}, and Adversarial NLI~\citep{,nie-etal-2020-adversarial}. We refer to the original works for more details.

\section{CREST Details} \label{sec:CREST_details}

\subsection{Masker}
For all datasets, the masker consists of a SPECTRA rationalizer that uses a T5-small encoder as the backbone for the encoder and predictor (see \S\ref{subsec:vanilla_rationalizer}). 
Our implementation is derived directly from its original source~\citep{guerreiro-martins-2021-spectra}.
We set the maximum sequence length to 512, truncating inputs when necessary. We employ a contiguity penalty of $10^{-4}$ for IMDB and $10^{-2}$ for SNLI. 
We train all models for a minimum of 3 epochs and maximum of 15 epochs along with early stopping with a patience of 5 epochs. We use AdamW~\citep{loshchilov2018decoupled} with a learning rate of $10^{-4}$ and weight decay of $10^{-6}$. 

\subsection{Editor}
For all datasets, CREST and MiCE editors consist of a full T5-small model~\citep{raffel2020exploring}, which includes both the encoder and the decoder modules. We use the T5 implementation available in the \textit{transformers} library~\citep{wolf-etal-2020-transformers} for our editor. We train all models for a minimum of 3 epochs and maximum of 20 epochs along with early stopping with a patience of 5 epochs. We use AdamW~\citep{loshchilov2018decoupled} with a learning rate of $10^{-4}$ and weight decay of $10^{-6}$. 
For both CREST and MiCE, we generate counterfactuals with beam search with a beam of size 15 and disabling bigram repetitions. 
We post-process the output of the editor to trim spaces and repetitions of special symbols (e.g., \texttt{</s>}).

\subsection{SPECTRA rationalizers} \label{subsec:CREST_details_SPECTRA_training}
All of our SPECTRA rationalizers share the same setup and training hyperparameters as the one used by the masker in \S\ref{sec:counterfactual_generation}, but were trained with distinct random seeds. 
We tuned the counterfactual loss weight $\alpha$ within $\{1.0, 0.1, 0.01, 0.001, 0.0001\}$, and $\lambda$ within $\{1.0, 0.1, 0.01, 0.001\}$ for models trained with agreement rationalization. 
More specifically, we performed hyperparameter tuning on the validation set, with the goal of maximizing in-domain accuracy. 
As a result, we obtained $\alpha = 0.01$ and $\lambda = 0.001$ for IMDB, and $\alpha = 0.01$ and $\lambda = 0.1$ for SNLI.

\section{Validity vs. Closeness}\label{sec:validity_x_closeness}

\begin{figure}[t]
    \centering
    \includegraphics[width=\columnwidth]{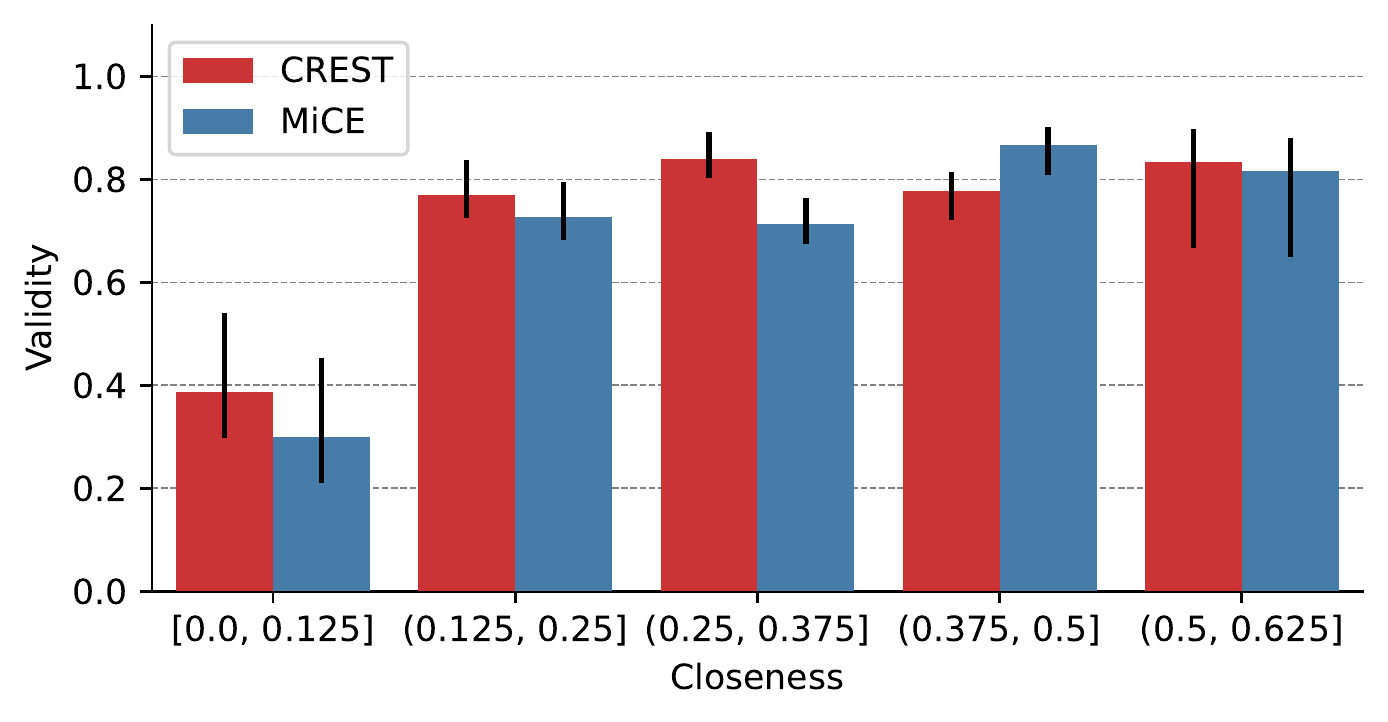} 
    \vspace{-0.7cm}
    \caption{Validity by binned closeness ranges for MiCE (30\% masking) and CREST (30\% masking). At lower closeness ranges, CREST produces more valid counterfactuals than does MiCE.}
    \label{fig:closeness_x_validity}
\end{figure}

To better assess the performance of CREST and MiCE by varying closeness, we plot in Figure~\ref{fig:closeness_x_validity} binned-validity scores of CREST and MiCE with 30\% masking on the revised SNLI dataset. Although CREST is deemed less valid than MiCE overall (\textit{cf.} Table~\ref{tab:results_counterfactuals_methods}), 
we note that CREST generates more valid counterfactuals in lower minimality ranges. 
This provides further evidence that CREST remains superior to MiCE on closeness intervals of particular interest for generating counterfactuals in an automatic way.

\section{Human Annotation}\label{sec:human_study}

The annotation task was conducted by four distinct individuals, all of whom are English-fluent PhD students. Two annotators were employed for IMDB and two for SNLI. 
The annotators were not given any information regarding the methods used to create each counterfactual, and the documents were presented in a random order to maintain source anonymity.
The annotators were presented with the reference text and its corresponding gold label. Subsequently, for each method, they were asked to assess both the validity and the naturalness of the resulting counterfactuals using a 5-point Likert scale.
We provided a guide page to calibrate the annotators' understating of validity and naturalness prior the annotation process. 
We presented hypothetical examples with different levels of validity and naturalness and provided the following instructions regarding both aspects: 
\begin{itemize}
    \item ``If every phrase in the text unequivocally suggests a counterfactual label, the example is deemed fully valid and should receive a top score of 5/5.''

    \item ``If the counterfactual text aligns with the style, tone, and grammar of real-world examples, it’s considered highly natural and deserves a score of 5/5.``
\end{itemize}

We measure inter-annotator agreement with a normalized and inverted Mean Absolute Difference (MAD), which computes a ``soft'' accuracy by averaging absolute difference ratings and normalizing them to a 0-1 range. We present the annotation results in Table~\ref{tab:annotation_stats}.
Our results show that humans agreed more on manual examples than on automatic approaches. 
On the other hand, for SNLI, annotators assigned similar scores across all methods.
In terms of overall metrics, including validity, naturalness, and agreement, the scores were lower for IMDB than for SNLI, highlighting the difficulty associated with the generation of counterfactuals for long movie reviews.

\begin{table}[t]
    \centering
    \small
    \setlength{\tabcolsep}{4pt}
    \begin{tabular}{lccc c ccc}
         \toprule
         & \multicolumn{3}{c}{\bf IMDB} & & \multicolumn{3}{c}{\bf SNLI} \\ 
         \cmidrule{2-4} 
         \cmidrule{6-8} 
         \bf Method & $v$ & $n$ & $r_o$ & & $v$ & $n$ & $r_o$ \\
         \midrule
         Manual   & 4.60 & 4.36 & 0.83   & & 4.89 & 4.90 & 0.95 \\
         MiCE     & 2.76 & 2.29 & 0.71   & & 4.35 & 4.71 & 0.94 \\
         CREST    & 4.06 & 3.44 & 0.76   & & 4.89 & 4.89 & 0.96 \\
         \cdashlinelr{1-8}
         \textit{Overall}  & 3.81 & 3.36 & 0.77   & & 4.71 & 4.83 & 0.95 \\
         \bottomrule
    \end{tabular}
    \caption{Annotation statistics. $v$ and $n$ represent the averaged validity and naturalness scores, whereas $r_o$ is the relative observed agreement computed with a normalized and inverted MAD.}
    \label{tab:annotation_stats}
\end{table}

\begin{figure*}[t]
    \centering
    \includegraphics[width=\textwidth]{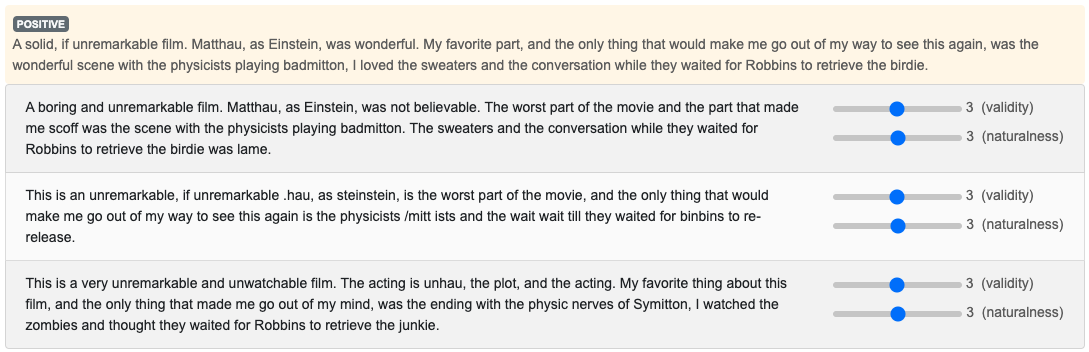}
    \caption{Snapshot of the annotation interface.}
    \label{fig:interface_snapshot}
\end{figure*}

\paragraph{Annotation interface.} Figure~\ref{fig:interface_snapshot} shows a snapshot of the interface used for the annotation, which is publicly available at \url{https://www.github.com/mtreviso/TextRankerJS}. 

\section{Counterfactual Data Augmentation Analysis}~\label{sec:cf_data_aug_analysis}

Previous studies on counterfactual data augmentation have found that model performance highly depends on the number and diversity of augmented samples~\citep{huang-etal-2020-counterfactually,joshi-he-2022-investigation}. To account for this, we investigate the effect of adding increasingly larger portions of CREST counterfactuals for data augmentation on the IMDB dataset.
Our findings are summarized in Table~\ref{tab:results_rationalizers_cf_data_aug}.

\begin{table}[t]
    \small
    \centering
    \resizebox{\columnwidth}{!}{%
    \setlength{\tabcolsep}{4pt}
    \begin{tabular}{l@{}rcccc}
        \toprule
        \bf Setup & \bf Data size & \bf RotTom & \bf SST-2 & \bf Amazon & \bf Yelp \\
        \midrule
        $F$ & 100\%           & 76.5 \scriptsize{$\pm$ 1.6}  & 79.8 \scriptsize{$\pm$ 1.6}  & 86.0 \scriptsize{$\pm$ 0.7}  & 88.5 \scriptsize{$\pm$ 0.7} \\
        \cdashlinelr{1-6}
        \multicolumn{3}{l}{\textit{With data augmentation:}} \\
        $F + C_H$ & +8\%      & 76.6 \scriptsize{$\pm$ 1.5}  & 80.7 \scriptsize{$\pm$ 1.3}  & 86.3 \scriptsize{$\pm$ 1.0}  & 89.1 \scriptsize{$\pm$ 1.2} \\
        $F + C_{S,V}$ & +1\%  & 77.2 \scriptsize{$\pm$ 1.1}  & 80.5 \scriptsize{$\pm$ 0.5}  & 86.1 \scriptsize{$\pm$ 0.2}  & 88.8 \scriptsize{$\pm$ 0.3}  \\
        $F + C_{S,V}$ & +2\%  & 76.2 \scriptsize{$\pm$ 1.2}  & \underline{80.8} \scriptsize{$\pm$ 0.8}  & 86.7 \scriptsize{$\pm$ 0.5}  & \underline{89.6} \scriptsize{$\pm$ 0.5}  \\
        $F + C_{S,V}$ & +4\%  & \textbf{77.7} \scriptsize{$\pm$ 0.8}  & \underline{80.8} \scriptsize{$\pm$ 0.7}  & \textbf{87.0} \scriptsize{$\pm$ 0.6}  & \textbf{89.8} \scriptsize{$\pm$ 0.6}  \\
        $F + C_{S,V}$ & +8\%  & 76.6 \scriptsize{$\pm$ 2.2}  & 80.2 \scriptsize{$\pm$ 1.7}  & 86.1 \scriptsize{$\pm$ 0.9}  & 88.2 \scriptsize{$\pm$ 1.0}  \\
        $F + C_{S,V}$ & +85\%  & 76.8 \scriptsize{$\pm$ 0.9} & 79.3 \scriptsize{$\pm$ 0.3} & 85.2 \scriptsize{$\pm$ 0.9} & 88.0 \scriptsize{$\pm$ 1.0}  \\
        $F + C_{S}$ & +100\%   & 76.7 \scriptsize{$\pm$ 1.0} & 80.6 \scriptsize{$\pm$ 0.6} & 86.4 \scriptsize{$\pm$ 0.6} & 89.1 \scriptsize{$\pm$ 0.5} \\
        \cdashlinelr{1-6}
        \multicolumn{3}{l}{\textit{With agreement regularization:}} \\
        $F\ \&\ C_{S, V}$ & 85\%    & 76.3 \scriptsize{$\pm$ 1.4} & 80.2 \scriptsize{$\pm$ 1.3} & 86.3 \scriptsize{$\pm$ 0.7} & 88.9 \scriptsize{$\pm$ 0.7}  \\
        $F\ \&\ C_{S}$ & 100\%      & \underline{77.3} \scriptsize{$\pm$ 2.3} & \textbf{81.1} \scriptsize{$\pm$ 2.4} & \underline{86.8} \scriptsize{$\pm$ 0.8} & 89.3 \scriptsize{$\pm$ 0.7}  \\
        \bottomrule
    \end{tabular}
    }
    \caption{OOD accuracy of SPECTRA rationalizers with different portions of augmented counterfactuals. \textbf{Bold}: top results; \underline{underlined}: second-best.}
    \label{tab:results_rationalizers_cf_data_aug}
\end{table}

\paragraph{Discussion.} 
We find that incorporating human-crafted counterfactuals ($F + C_H$) improves SPECTRA performance on all OOD datasets. On top of that, we note that using a small proportion (4\% of the full IMDB) of valid CREST counterfactuals ($F + C_{S,V}$) through data augmentation also leads to improvements on all datasets and outweighs the benefits of manual counterfactuals. 
This finding confirms that, as found by PolyJuice~\citep{wu-etal-2021-polyjuice}, synthetic counterfactuals can improve model robustness.
Conversely, as the number of augmented counterfactuals increases ($85\%$), the performance on OOD datasets starts to decrease, which is also consistent with the findings of \citet{huang-etal-2020-counterfactually}. 
When augmenting the entire training set we obtain an increase of accuracy, suggesting that the counterfactual loss weight ($\alpha$) was properly adjusted on the validation set.
Finally, we observe that while applying \crestrat only on valid examples ($F\ \& \ C_{S,V}$) degrades performance, applying \crestrat on all paired examples ($F\ \& \ C_{S}$) maintains a high accuracy on OOD datasets and concurrently leads to strong results on in-domain and contrast datasets~(see Table~\ref{tab:results_rationalizers_in_domain}).

\section{Computing infrastructure}

Our infrastructure consists of four machines with the specifications shown in Table~\ref{table:computing_infrastructure}. 
The machines were used interchangeably and all experiments were carried in a single GPU. 

\begin{table}[!htb]
    \small
    \centering
    \begin{tabular}{ll}
        \toprule
        \sc GPU & \sc CPU  \\
        \midrule
        4 $\times$ Titan Xp - 12GB           & 16 $\times$ AMD Ryzen - 128GB \\
        4 $\times$ GTX 1080Ti - 12GB        & 8 $\times$ Intel i7 - 128GB \\
        3 $\times$ RTX 2080Ti - 12GB        & 12 $\times$ AMD Ryzen - 128GB \\
        3 $\times$ RTX 2080Ti - 12GB        & 12 $\times$ AMD Ryzen - 128GB \\
        \bottomrule
    \end{tabular}
    \caption{Computing infrastructure.} 
    \label{table:computing_infrastructure}
\end{table}

\section{Examples of Counterfactuals}
\label{app:examples_counterfactuals}

Table~\ref{tab:examples_counterfactuals} shows examples of counterfactuals produced by MiCE and CREST with 30\% masking.

\begin{table*}[!htb]
    \centering
    \small
    \fontsize{8}{9}\selectfont
    \begin{tabular}{r@{\ \ }m{0.9\textwidth}}
    \toprule
    \multicolumn{2}{c}{\textit{Sentiment Classification:}} \\
    \vspace{0.1cm}
    \textbf{Input:} & If you haven't seen this, it's terrible. It is pure trash. I saw this about 17 years ago, and I'm still screwed up from it.
    \\
    \vspace{0.1cm}
    \textbf{MiCE:} & If you haven't seen this, it's a great movie. I saw this about 17 years ago, and I'm still screwed up from it.
    \\
    \textbf{CREST:} & If you haven't seen this movie, it's worth seeing. It's very funny. I saw it about 17 years ago, and I'm still screwed up from it.
    \\
    \cdashlinelr{1-2}
    \vspace{0.1cm}
    \textbf{Input:} & Touching; Well directed autobiography of a talented young director/producer. A love story with Rabin's assassination in the background. Worth seeing !
    \\
    \vspace{0.1cm}
    \textbf{MiCE:} & Watching abiography of a very young writer/producer. A great story of Rabin's assassination in the background! Worth seeing!!
    \\
    \textbf{CREST:} & This is the worst film of a talented young director/producer. And Rabin's assassination in the background is even worse!
    \\
    \cdashlinelr{1-2}
    \vspace{0.1cm}
    \textbf{Input:} & A solid, if unremarkable film. Matthau, as Einstein, was wonderful. My favorite part, and the only thing that would make me go out of my way to see this again, was the wonderful scene with the physicists playing badmitton, I loved the sweaters and the conversation while they waited for Robbins to retrieve the birdie.
    \\
    \vspace{0.1cm}
    \textbf{MiCE:} & This is an unremarkable, if unremarkable .hau, as steinstein, is the worst part of the movie, and the only thing that would make me go out of my way to see this again is the physicists /mitt ists and the wait wait till they waited for binbins to re-release.
    \\
    \textbf{CREST:} & This is a very unremarkable and unwatchable film. The acting is unhau, the plot, and the acting. My favorite thing about this film, and the only thing that made me go out of my mind, was the ending with the physic nerves of Symitton, I watched the zombies and thought they waited for Robbins to retrieve the junkie.
    \\
    \cdashlinelr{1-2}
    \vspace{0.1cm}
    \textbf{Input:} & I saw this film earlier today, and I was amazed at how accurate the dialog is for the main characters. It didn't feel like a film - it felt more like a documentary (the part I liked best). The leading ladies in this film seemed as real to me as any fifteen year-old girls I know. All in all, a very enjoyable film for those who enjoy independent films.
    \\
    \vspace{0.1cm}
    \textbf{MiCE:} & I saw this film earlier today, and I was amazed at how bad the film is for the sake of a film - it feels more like thanthe part I played in this film. To me - fifteen year-old s I don't know. All in all this is a bad film for those who like independent films :
    \\
    \textbf{CREST:} & I saw this movie earlier today, and I was surprised at how bad it is for the first time. It's not a good movie - it's just a bad movie (the only thing I can say about it). The acting is awful to me as any fifteen year-old as I can. All in all, the movie is a waste of time for me.
    \\
    \midrule
    \multicolumn{2}{c}{\textit{Natural Language Inference:}} \\
    \textbf{Prem:} & A large group of people walking in a busy city at night. \\
    \vspace{0.1cm}
    \textbf{Hyp:}   & People are outside in a park.
    \\
    \vspace{0.1cm}
    \textbf{MiCE:} & People are walking in a city at night
    \\
    \textbf{CREST:} & People walking in a city.
    \\
    \cdashlinelr{1-2}
    \textbf{Prem:} & Players from two opposing teams wearing colorful cleats struggle to gain control over a ball on an AstroTurf field. \\
    \vspace{0.1cm}
    \textbf{Hyp:}   & The players are playing a sport.
    \\
    \vspace{0.1cm}
    \textbf{MiCE:} &  The players are playing chess at home
    \\
    \textbf{CREST:} & The players are sitting on a couch.
    \\
    \cdashlinelr{1-2}
    \textbf{Prem:} & A woman is in the middle of hitting a tennis ball. \\
    \vspace{0.1cm}
    \textbf{Hyp:}   & A woman is playing tennis.
    \\
    \vspace{0.1cm}
    \textbf{MiCE:} &   A woman is playing basketball at home
    \\
    \textbf{CREST:} & A woman is playing basketball.
    \\
    \cdashlinelr{1-2}
    \textbf{Prem:} & Two boys with blond-hair, wearing striped shirts on a bed. \\
    \vspace{0.1cm}
    \textbf{Hyp:}   & Children playing in the park.
    \\
    \vspace{0.1cm}
    \textbf{MiCE:} &  Children are on the bed.
    \\
    \textbf{CREST:} & Boys are on the bed.
    \\
    \cdashlinelr{1-2}
    \textbf{Prem:} & Bubbles surround a statue in the middle of a street. \\
    \vspace{0.1cm}
    \textbf{Hyp:}   & There are bubbles around the statue.
    \\
    \vspace{0.1cm}
    \textbf{MiCE:} &  There are bubbles surround the statue.
    \\
    \textbf{CREST:} & Bubbles are in the ocean.
    \\
    \cdashlinelr{1-2}
    \textbf{Prem:} & A young girl is standing in a kitchen holding a green bib. \\
    \vspace{0.1cm}
    \textbf{Hyp:}   & A boy is playing with a firetruck.
    \\
    \vspace{0.1cm}
    \textbf{MiCE:} &  A child is in a fire place
    \\
    \textbf{CREST:} & A girl is holding a bib.
    \\
    \bottomrule
    \end{tabular}
    \caption{Examples of original inputs from the IMDB and SNLI datasets followed by synthetic counterfactuals produced by MiCE and CREST with 30\% masking. 
    }
    \label{tab:examples_counterfactuals}
\end{table*}

\end{document}